# Tracking The Untrackable:
# Learning to Track Multiple Cues with Long-Term Dependencies

Amir Sadeghian, Alexandre Alahi, Silvio Savarese
Stanford University
{amirabs,alahi,ssilvio}@cs.stanford.edu

## Abstract

*The majority of existing solutions to the Multi-Target Tracking (MTT) problem do not combine cues in a coherent end-to-end fashion over a long period of time. However, we present an online method that encodes long-term temporal dependencies across multiple cues. One key challenge of tracking methods is to accurately track occluded targets or those which share similar appearance properties with surrounding objects. To address this challenge, we present a structure of Recurrent Neural Networks (RNN) that jointly reasons on multiple cues over a temporal window. We are able to correct many data association errors and recover observations from an occluded state. We demonstrate the robustness of our data-driven approach by tracking multiple targets using their appearance, motion, and even interactions. Our method outperforms previous works on multiple publicly available datasets including the challenging MOT benchmark.*

## 1. Introduction

Architectures based on neural networks have become an essential instrument in solving perception tasks and have shown to approach human-level accuracy in classifying images [18, 19]. However, the status quo of the Multi-Target Tracking (MTT) problem is still far from matching human performance [61, 75]. This is mainly because it is difficult for neural networks to capture the inter-relation of targets in time and space using multi-modal cues (e.g., appearance, motion, and interactions). In this work, we tackle the MTT problem by jointly learning a representation that takes into account several cues over a time period in an end-to-end fashion (see Figure 1).

The objective of MTT is to infer trajectories of targets as they move around. It covers a wide range of applications such as sports analysis [43, 51, 76], biology (e.g., birds [44], ants [28], fish [66, 67, 15], cells [45, 39]), robot navigation [11, 12], and autonomous driving vehicles [13, 57].

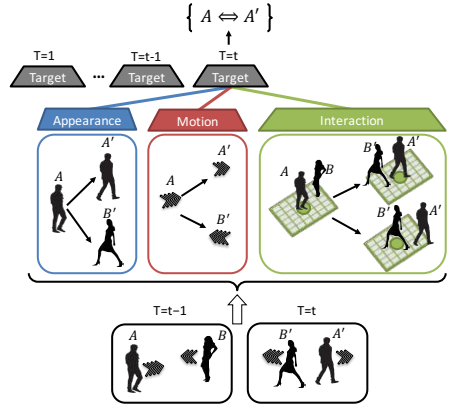

Figure 1. We present a method based on a structure of RNNs (each RNN is depicted by a trapezoid) that learns to encode long-term temporal dependencies across multiple cues (appearance, motion, and interaction). Our learned representation is used to compute the similarity scores of a "Tracking-by-detection" algorithm. [16]

We follow the "tracking-by-detection" paradigm whereby detection outputs are to be connected across video frames. This is often formulated as an optimization problem with respect to a graph [55, 56]. Each detection is represented by a node, and edges encode the similarity scores. Over the past decades, researchers have made significant progress in proposing techniques to solve the optimal assignments of graph-based formulations [86, 1, 34, 64]. However, their MTT performances are limited by the specific design choices of their representation and the corresponding similarity function.

In crowded environments, occlusions, noisy detections (e.g., false alarms, missing detections, non-accurate bounding), and appearance variability are very common. In traditional MTT approaches, representations and similarity functions are hand-crafted in an attempt to capture similar appearance and motion across adjacent temporal frames [34, 64, 75, 77]. In contrast, we propose a method to encode long-term temporal dependencies across multiple cues without the need to hand specify parameters or weights.

Our framework is based on a structure of Recurrent Neural Networks (RNN), which has also shown benefits in other applications [26]. The rest of the paper is as follows. In Section 3, we present details on the inputs of each RNN and learning a representation that can be used to compute a similarity score in an end-to-end fashion. Our appearance model is an RNN constructed on a Convolutional Neural Network (CNN) whose purpose is to classify if a detection is similar to a target instance at different time frames. Our motion and interaction models leverage two separate Long Short-Term Memory (LSTM) networks that track the motion and interactions of targets for longer period – suitable for presence of long-term occlusions. We then combine these networks into a structure of RNN to learn to reason jointly on different cues across time. Our method runs online without the need to see future frames. In Section 4, we present a detailed evaluation of our framework using multiple benchmarks such as the MOT challenge [37, 46] and Stanford drone dataset [59].

## 2. Related Work

In recent years, tracking has been successfully extended to scenarios with multiple targets [52, 38, 24, 75]. As opposed to single target tracking approaches which have been constructing a sophisticated appearance model to track a single target in different frames, multiple target tracking does not mainly focus on appearance model. Although appearance is an important cue, relying only on appearance can be problematic in MTT scenarios where the scene is highly crowded or when targets may share the same appearance. To this end, some works have been improving the appearance model [17, 7], while others have been combining the dynamics and interaction between targets with the target appearance [59, 3, 55, 77, 9, 62, 56].

### 2.1. Appearance Model

Simple appearance models are widely used in MTT. Many models are based on raw pixel template representation for simplicity [77, 4, 74, 55, 54], while color histogram is the most popular representation for appearance modeling in MTT approaches [9, 38, 68, 35]. Other approaches use covariance matrix representation, pixel comparison representation, SIFT-like features, or pose features [25, 83, 29, 22, 50]. Recently, deep neural network architectures have been used for modeling appearance [21, 36, 84]. In these architectures, high-level features are extracted by convolutional neural networks trained for a specific task. The appearance module of our model shares some characteristics with [21], but differs in two crucial ways: first, we handle occlusions and solve the re-identification task by learning a similarity metric between two targets. Second, the network architecture is different and we use a different loss function which we will describe in Section 3.2.

### 2.2. Motion Model

The target motion model describes how a target moves. The motion cue is a crucial cue for MTT, since knowing the likely position of targets in the future frames will reduce the search space and hence increases the appearance model accuracy. Popular motion models used in MTT are divided into linear and non-linear motion models. Linear motion models follow a linear movement with constant velocity across frames. This simple motion model is one of the most popular models in MTT [7, 47, 63, 82, 53]. However, there are many cases linear motion models can not deal with long-term occlusions; to remedy this, non-linear motion models are proposed to produce a more accurate prediction [78, 79, 10]. We present a Long Short-Term Memory (LSTM) model which learns to predict similar motion patterns. It is a fully data-driven approach that can handle noisy detections.

### 2.3. Interaction Model

Most tracking techniques assume that each target has an independent motion model. This simplification can be problematic in crowded scenes. Interaction models capture interactions and forces between different targets in a scene [20, 23, 73]. Two most popular interaction models are the social force models introduced by [20] and the crowd motion pattern model [23]. Social force models are also known as group models. In these models, each target reacts to energy potentials caused by interactions with other objects through forces (repulsion or attraction), while trying to keep a desired speed and motion direction [59, 3, 55, 77, 9, 62, 56]. Crowd motion pattern models are another type of interaction models used in MTT, inspired by the crowd simulation literature [85, 65]. In general, these kind of models are usually used for over-crowded scenes [90, 48, 32, 33, 60, 58]. The main drawback of most these methods is that they are limited to a few hand-designed force terms, such as collision avoidance or group attraction. Recently, Alahi et al. [3] proposed to use Long Short-Term Memory networks to jointly reason across multiple individuals (referred to as social LSTM). They presented an architecture to forecast the long-term trajectories of all targets. We use a similar LSTM based architecture. However, our data-driven interaction model is trained to solve the re-identification task as opposed to the long-term prediction.

Finally, when reasoning with multiple cues, previous works combine them in a hand-crafted fashion without adequately modeling long-term dependencies. None of the previous method discussed in sections 2.1, 2.2, and 2.3 combine appearance, motion, and interaction cues in a coherent end-to-end architecture. In this work, we propose a structure of RNNs to cope with such limitations of previous works. We learn a representation that encodes long-term temporal dependencies across multiple cues, i.e., appearance, motion, and interaction automatically in a data-driven fashion.

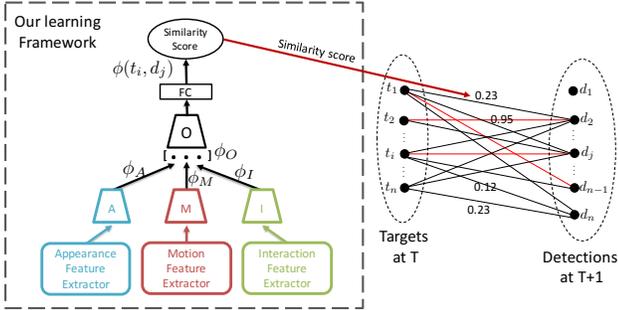

Figure 2. We use a structure of RNNs (the dashed rectangle) to compute the similarity scores between targets $t_i$ and detections $d_j$. The scores are used to construct a bipartite graph between the targets and detections. The structure of RNNs is comprised of three RNNs – Appearance (A), Motion (M), and Interaction (I) – that are combined through another RNN (referred to as the target RNN (O)).

## 3. Multi-Target Tracking Framework

The task of Multi-Target Tracking (MTT) consists of detecting multiple targets at each time frame and matching their identities in different frames, yielding to a set of target trajectories over time. We address this problem by using a "tracking-by-detection" paradigm. As the input, the detection results are produced by an object detector. Given a new frame, the tracker computes the similarity scores between the already tracked targets and the newly detected objects (more details in section 3.1). These similarity scores are calculated using our framework (as described in Figure 2). They are used to connect the detections $d_j$ and targets $t_i$ in a bipartite graph, as shown in right-side of Figure 2. Then, the Hungarian algorithm [49] is used to find the optimal assignments. In this work, we propose a new method to compute these similarity scores.

### 3.1. Overall Architecture

We have identified appearance cues, motion priors, and interactive forces as critical cues of the MTT problem. As discussed in the introduction, combining these cues linearly is not necessarily the best way to compute the similarity score. We instead propose to use a structure of RNNs to combine these cues in a principled way.

In our framework, we represent each cue with an RNN. We refer to the RNNs obtained from these cues as appearance (A), motion (M), and interaction (I) RNNs. The features represented by these RNNs ($\phi_A$, $\phi_M$, $\phi_I$) are combined through another RNN which is referred to as target (O)

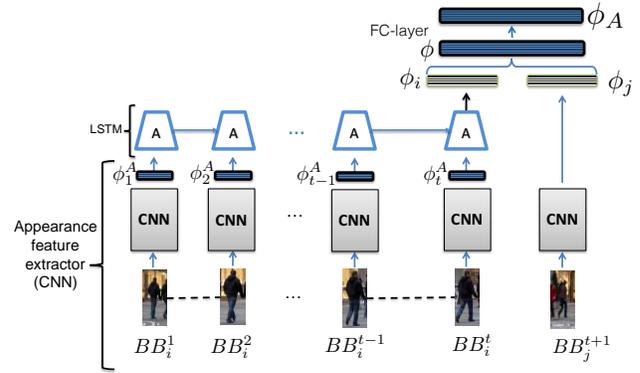

Figure 3. Our appearance model. The inputs are the bounding boxes of target $i$ from time 1 to $t$, and detection $j$ at time $t+1$ we wish to compare. The output is a feature vector $\phi_A$ that encodes if the bounding box at time $t+1$ corresponds to a specific target $i$ at time $1, 2, \ldots, t$. We use a CNN for our appearance feature extractor.

RNN. More details on the architecture and training process of these RNNs can be found in sections 3.2, 3.3, 3.4, and 3.5 respectively. The target RNN outputs a feature vector, $\phi(t, d)$, which is used to output the similarity between a target $t$ and a detection $d$.

By using RNNs, more precisely LSTM networks, we have the capacity to encode long-term dependencies in the sequence of observations. Traditionally, similarity scores in a graph-based tracking framework were computed given only the observation from the previous frame, i.e., a pairwise similarity score [86, 1, 34, 64]. Our proposed similarity score is computed by reasoning on the sequence of observations. In Section 4.2, we demonstrate the power of our representation by reasoning on a sequence of variable length as opposed to a pairwise similarity score. In the rest of this section, we describe each component of our method.

### 3.2. Appearance

The underlying idea of our appearance model is that we can compute the similarity score between a target and candidate detection based on purely visual cues. More specifically, we can treat this problem as a specific instance of *re-identification*, where the goal is to take pairs of bounding boxes and determine if their content corresponds to the same target. Therefore, our appearance model should subtle similarities between input pairs, as well as be robust to occlusions and other visual disturbances. The appearance model's output feature vector is produced by an RNN (A), which in turn receives its input from the appearance feature extractor (see Figure 3).

**Architecture:** Our appearance RNN (A) is an LSTM that accepts as inputs the appearance features from the appearance feature extractor ($\phi_1^A, \ldots, \phi_t^A$) and produces $H$- dimensional output $\phi_i$ for each timestep. The appearance features are the last hidden layer features of a Convolutional Neural Network (CNN).

Let $BB_i^1, \ldots, BB_i^t$ be the bounding boxes of target $i$ at timesteps $1, \ldots, t$ and $BB_j^{t+1}$ be the detection $j$ we wish to compare with target $i$. The CNN accepts the raw content within each bounding box and passes it through its layers until it finally produces a 500-dimensional feature vector ($\phi_i^A$). We also pass $BB_j^{t+1}$ (which we wish to determine whether it corresponds to the true appearance trajectory of target $i$ or not) through the same CNN that maps it to an $H$-dimensional vector $\phi_j$. The LSTM's output $\phi_i$ is then concatenated with this vector, and the result $\phi$ is passed to another FC layer which brings the $2H$ dimensional vector to a $k$ dimensional feature vector $\phi_A$ (as illustrated in Figure 3). We pre-train our appearance model using a Softmax classifier for $0/1$ classification problem, whether $BB_j^{t+1}$ corresponds to a true appearance trajectory $BB_i^1, \ldots, BB_i^t$. When combining with other cues, we use $\phi_A$ of size 500 as part of the input to our target RNN (O).

Note that we use a 16-layer VGGNet as our CNN in Figure 3. We begin with the pre-trained weights of this network, remove the last FC layer and add an FC layer of size 500 so that the network now outputs a 500-dimensional vector. We then train this CNN for the re-identification task for which the details can be found in Section 4.3.

### 3.3. Motion

The second cue of our overall framework is the independent motion property of each target. It can help tracking targets that are occluded or lost. One key challenge is to handle the noisy detections. Even when the real motion of a target is linear, since detections can be noisy, the sequence of coordinates hence velocities can be non-linear – especially if we reason on the image plane. We train a Long Short-Term Memory (LSTM) network on trajectories of noisy 2D velocities (extracted by our motion feature extractor) to be able to learn this non-linearities from data (see figure 4).

**Architecture:** Let the velocity of target $i$ at the $t$-th timestep be defined as:
$v_i^t = (vx_i^t, vy_i^t) = (x_i^t - x_i^{t-1}, y_i^t - y_i^{t-1})$,
where $(x_i^t, y_i^t)$ are the 2D coordinates of each target on the image plane (center of the bounding boxes).

Our motion RNN (M) is an LSTM that accepts as inputs the velocities of a specific target at timesteps $1, \ldots, t$ as motion features, and produces an $H$-dimensional output $\phi_i$. We also pass the velocity vector of the detection $j$ at timestep $t+1$ (which we wish to determine whether it corresponds to the true trajectory of target $i$ or not) through a fully-connected layer that maps it to an $H$-dimensional vector $\phi_j$ (this makes $\phi_j$ the same size as $\phi_i$). The LSTM output is then concatenated with this vector, and the result is passed to another fully connected layer which brings the $2H$ dimensional vector to a $k$ dimensional feature vector $\phi_M$ (as illustrated in Figure 4). We pre-train our motion model using a Softmax classifier for $0/1$ classification problem, whether velocity $v_j^{t+1}$ corresponds to a true trajectory $v_i^1, \ldots, v_i^t$. When combining with other cues, we use $\phi_M$ of size 500 as part of the input to our target RNN (O).

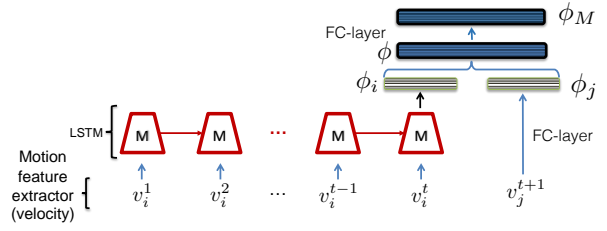

Figure 4. Our motion model. The inputs are the 2D velocities of the target (on the image plane). The output is a feature vector $\phi_M$ that encodes if velocity $v_j^{t+1}$ corresponds to a true trajectory $v_i^1, v_i^2, \ldots, v_i^t$.

### 3.4. Interaction Model

The motion of a particular target is governed not only by its own previous motion, but also by the behavior of nearby targets. We incorporate this cue into our overall framework by formulating an interaction model. Since the number of nearby targets can vary, in order to use the same size input, we model the neighborhood of each target as a fixed size occupancy grid. The occupancy grids are extracted from our interaction feature extractor. For each target, we use an LSTM network to model the sequence of occupancy grids (see Figure 6).

**Architecture:** Let $O_i^1, O_i^2, \ldots, O_i^t$ represent the 2D occupancy grid for target $i$ at timesteps $1, \ldots, t$. The positions of all the neighbors are pooled in this map. The m, n element of the map is simply given by:

$$O_i^t(m, n) = \vee_{j \in \mathcal{N}_i} \mathbf{1}_{mn}[x_j^t - x_i^t, y_j^t - y_i^t]$$

Where $\vee$ is logical disjunction, $\mathbf{1}_{mn}[x, y]$ is an indicator function to check if the person located at $(x, y)$ is in the (m, n) cell of the grid, and $\mathcal{N}_i$ is the set of neighbors corresponding to person $i$. The map is further represented as a vector (see Figure 6). Note that all the 2D locations of targets are their equivalent bounding box centers on the image plane.

Our interaction RNN (I) is an LSTM that accepts as input the occupancy grids centered on a specific target for timesteps $1, \ldots, t$ (extracted by the interaction feature extractor) and produces $H$-dimensional output $\phi_i$ for each timestep. We also pass the occupancy grid of detection $j$ at timestep $t+1$ (which we wish to determine whether it corresponds to the true trajectory of target $i$ or not) through a fully-connected layer that maps it to an $H$-dimensional vector space $\phi_j$ (this makes $\phi_j$ the same size as $\phi_i$). The LSTM output is then concatenated with this vector resulting into vector $\phi$, which is passed to another fully connected layer that brings the $2H$ dimensional vector $\phi$ to the space of $k$ dimensional feature vector $\phi_I$ (as illustrated in Figure 5). We pre-train our interaction model using a Softmax classifier for 0/1 classification problem. Similar to motion model, when combining with other cues, we use $\phi_I$ of size 500 as part of input to our target RNN (O).

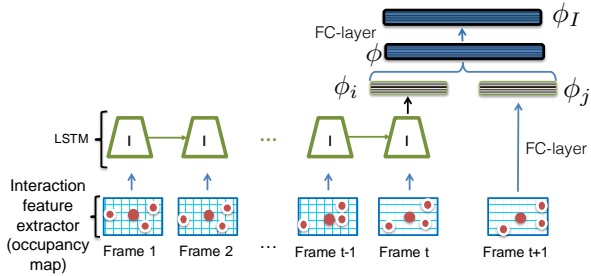

Figure 5. Our interaction model. The inputs are the occupancy maps across time (on the image plane). The output is a feature vector $\phi_I$ that encodes if the occupancy map at time $t+1$ corresponds to a true trajectory of occupancy maps at time $1, 2, \ldots, t$.

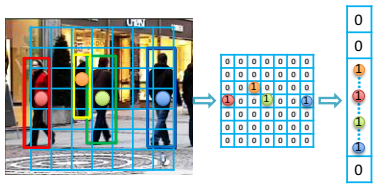

Figure 6. Illustration of the steps involved in computing the occupancy map. The location of the bounding box centers of nearby targets are encoded in a grid –occupancy map– centered around the target. For implementation purposes, the map is represented as a vector.

### 3.5. Target

Our overarching model shown in figure 2 is constructed by combining the appearance, motion, and interaction RNNs through another RNN which is referred to as target RNN (O).

The training proceeds in two stages:

(i) First, networks A, M and I (corresponding to appearance, motion, and interaction RNNs) as well as the CNN (appearance feature extractor) are pre-trained separately. We use a standard Softmax classifier and cross-entropy loss. Each RNN outputs the probabilities for the positive and negative classes, where positive indicates that the new detected object matches the previous trajectory of the target (in either case of appearance, motion, or interaction properties, depending on the RNN in charge), and negative indicates otherwise.

| Method | MOTA | MOTP | Rcll | Prcn | MT | ML |
|---|---|---|---|---|---|---|
| MDP [75] + Lin | 51.5 | 74.2 | 74.1 | 80.1 | 44.2% | 20.9% |
| MDP + SF [77] | 73.5 | 77.1 | 84.4 | 91.5 | 58.1% | 25.5% |
| MDP + SF-mc [59] | 75.6 | 78.2 | 86.1 | 92.6 | 60% | 23.2% |
| **Ours (MOT)** | 78.6 | 79.4 | 88.2 | 93.9 | 69.7% | 19.5% |
| **Ours (MOT+Drone)** | 82.9 | 80.3 | 92.3 | 95.3 | 85% | 15.2% |

Table 1. MOT tracking results on Stanford Drone Dataset. Our (MOT) version has been only trained on the MOT challenge training data and has not been fine-tuned on the Stanford drone dataset. Whereas, our (MOT+Drone) version has been also fine-tuned on the drone dataset.

(ii) Second, the target RNN is jointly trained end-to-end with the component RNNs A, M, and I. The output vectors of the A, M, and I networks are concatenated into a single feature vector and serve as input to the target RNN. Our target RNN has the capacity to learn long term dependencies of all cues across time. The last hidden state of the target RNN ($H$ dimensional) goes through a fully-connected layer resulting in a feature vector $\phi(t, d)$ that encodes all these long term dependencies of all cues across time. Our target RNN is also trained to perform the task of data association – outputs the score of whether a detection (d) corresponds to a target (t) from $\phi(t, d)$ using a Softmax classifier and cross-entropy loss.

In both above training stages, the networks are trained using MOT15 and MOT16 training data, in which positive examples are true pedestrian trajectories (consisting of appearances, velocities, and occupancy maps depending of the RNN), and negative examples are constructed by altering the pedestrian's appearance or location in the final frame of the trajectory simply by choosing another target's properties for the final frame.

## 4. Experimental Results

We have presented our multi-cue representation learning framework to compute the similarity score between a sequence of observations and a new detection. We use our learned representation to tackle the Multi-target Tracking problem. We first present the overall performance of our framework on the MOT challenge [37] and then present more insights and analysis on our representation.

### 4.1. Multi-Target Tracking

To recall, we use our learned representation in the MDP framework [75]. We have one target LSTM for each target, and the MDP framework tracks the targets using the similarity computed with our learned representation.

**Metrics.** We report the same metric as the suggested ones in the MOT2D Benchmark challenge [37]: Tracking Accuracy (MOTA), Multiple Object Tracking Precision (MOTP), Mostly Track targets (MT), Mostly Lost tar-gets (ML), False Positives (FP), False Negatives (FN), ID Switches (IDS), and finally the number of frames processed in one second (Hz) which indicates the speed of a tracking method.

**Implementation Details.** In all experiments the values of parameters H, k, and sequence lengths are 128, 100, and 6 respectively for all RNNs. Moreover, in section 3.4 the image is sampled uniformly with a 15*15 grid where a 7*7 sub-grid centered around a specific person is used as its occupancy grid. The network hyper-parameters are chosen by cross validation and our framework was trained with Adam update. Training the RNN's occurs from scratch, with mini-batch size of 64, and learning rate of 0.002, sequentially decreased every 10 epochs by a factor of 10 (for 50 epochs). Note that this is same for training all RNNs. Moreover, we use our method in the MDP framework [75]. For each target, MDP has two processes. First, it independently tracks the target with a single object tracker based on optical flow. Then, when the target gets occluded, the single object tracker stop tracking and a bi-partite graph is constructed similar to Figure 2. The Hungarian algorithm is used to recover occluded targets. Note that MDP also proposes to learn a similarity score given a hand-crafted representation. We replace their representation with the output of our target RNN ($\phi(t, d)$) to demonstrate the strength of our learning method.

**MOT Challenge Benchmark.** We report the quantitative results of our method on the 2DMOT 2015 Benchmark [37], and MOT16 [46] in Table 5 and 6. This challenges share the training and testing set for 11 and 14 sequences respectively. We used their publicly shared noisy detections. Our method outperforms previous methods on multiple metrics such as the MOTA, MT, and ML. Our MOTA even outperforms offline methods (in 2015 challenge) that have access to the whole set of future detections to reason on the data association step. Using long term dependencies of multiple cues makes our method to recover back to the right target after an occlusion or drift; hence we have higher MT and lower ML but our IDS is higher. Indeed, when targets are occluded, our method can wrongly assign them to other detections. But when the targets re-appear, our method re-match them with the correct detections. Such process leads to a high number of switches. Nevertheless, the MT metric remains high.

The impact of our learned representation becomes evident compared with the previously published MDP method. By only switching the representation and keeping the same data association method proposed in [75], we obtain a 20% relative boost in MOTA. The benefits of our representation are further emphasized with the Stanford dataset [59].

**Stanford Drone Dataset.** As we have mentioned before, one of the main advantages of our model compared to other multi-target tracking methods is the similarity score which is a function of multiple cues across time and seeks to use the right cues at each time. Often, some cues should vote for the similarity score since the others are not discriminant enough or very noisy. To test the power of our method, we also conduct experiments by testing our multi-target tracking experiments on videos that are very different from the MOT challenge [37], i.e., the Stanford Drone Dataset [59]. All targets are small and hence appearance models might be faulty (as illustrated in Figure 9). In table 1, we compare our method with previously reported MDP-based methods. Our method outperforms all the MDP-based methods on all metrics. Even without fine-tuning our representation on the drone dataset, our method outperforms previous works. After fine-tuning, we obtain the best performance as expected. It shows the power of a data-driven method to learn a representation over any input signal.

In the reminder of this section, we analyze the performance of our representation with an ablation study as well as more insights on our appearance on more specific tasks.

### 4.2. Ablation Study

The underlying motivation of our proposed framework is to address the following two challenges (as listed in the introduction): effectively modeling the history of each cue, and effectively combining multiple cues. We now present experiments towards these two goals on the validation set of the 2DMOT2015 challenge [37]. We use the same evaluation protocol (training and test splits) as in [75] for our validation set.

**Impact of the History.** One of the advantages of our representation compared to the previous ones is the capacity to learn long term dependencies of cues across time, i.e., retaining information from the past. We investigate the impact of changing the sequence length of the LSTMs on tracking accuracy, where sequence length of an LSTM is the number of unrolled time steps used while training the LSTM. Figure 7 (b) shows the MOTA score of different components for the validation set, under different LSTM sequence lengths for our target LSTM. We can see that increasing the LSTM sequence length positively impacts the MOTA. The performance saturates after 3 frames on the Stanford drone dataset and after 6 frames on the MOTChallenge dataset. These results confirm our claim that RNN can effectively model the history of a cue. Moreover, the difference between the MOT and Stanford dataset can be explained by the difference in the datasets. The drone dataset does not have any long term occlusion whereas the MOT has full long-term occlusions. Our framework learns to encode long-term temporal dependencies across multiple cues that helps recovering from long-term occlusions. We claim if most occlusions are less than $n$ frame long we at least need to keep dependencies over past $n$ frames to be able to recover the object from an occlusion. Figure 7 (a) depicts that most occlusions (more than 80 percent) happen for less than 6 frames in the MOT dataset which supports why the MOTA saturates after sequence length of 6, see figure 7 (b). Whereas since the Drone dataset does not have any long term occlusions our model does not need long term dependencies. Nevertheless, we can see that modeling the sequence of observations on both datasets positively impacts the similarity score hence tracking performance.

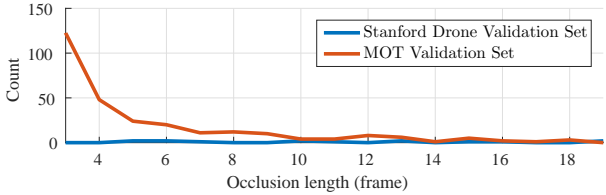

(a)

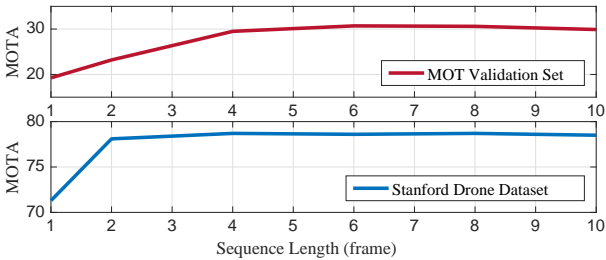

(b)

Figure 7. (a) Occlusion length distribution in MOT and Stanford Drone dataset validation set. (b) Analysis of the used sequence length (memory) for our model on the MOT validation set for both datasets. We report the MOTA scores.

| Tracker | MOTA | MOTP | MT | ML | FP | FN | IDS |
|---|---|---|---|---|---|---|---|
| Ours | **30.8** | **73.8** | **14** | **51.7** | **2,563** | **13,127** | **98** |
| Exp 1 | 18.2 | 71.2 | 7.1 | 72.1 | 3,851 | 15,893 | 350 |
| Exp 2 | 12.9 | 71.0 | 4.3 | 75.9 | 4,259 | 16,751 | 396 |

Table 2. Analysis of our model on the MOT validation set compared to FC baselines. First, using FC only instead of the target RNN (O) or top RNN (Exp 1) and second, FC layer for all RNNs (Exp 2).

In order to further support our use of RNNs for modeling the temporal dependencies in the MTT framework we conduct experiments using fully-connected layers instead of RNNs. We provide results of two experiments, one replacing only the target LSTM with an FC, and a second experiment in which we replaced all LSTM networks with FCs. Table 2 shows the results of this experiment.

**Impact of Multiple Cues.** We investigate the contribution of different cues in our framework by measuring the performance in terms of MOTA on the validation set. Figure 8 presents the results of our ablation study. The appearance cue is the most important one. Each cues helps to increase the performance. It is worth to point out our proposed interaction cue positively impacts the overall performance. Our proposed target LSTM (in charge of combining all the other RNNs) effectively reason on all the cues to increase the performance. Table 3 reports more details on the impact of each cue on the various tracking metrics.

| Tracker | MOTA | MOTP | MT | ML | FP | FN | IDS |
|---|---|---|---|---|---|---|---|
| A+M+I | 30.8 | 73.8 | **14** | **51.7** | **2,563** | **13,127** | **98** |
| A+M | 28.8 | **73.9** | 13.5 | 52.1 | 2,776 | 13,361 | 134 |
| A+I | 27.4 | 73.9 | 12 | 53.4 | 2,679 | 13,991 | 136 |
| M+I | 22 | 73.8 | 9.8 | 52.1 | 2,714 | 14,954 | 298 |
| A | 23.7 | 73.7 | 11.5 | 55.6 | 3,359 | 14,001 | 138 |
| M | 19.2 | 73.7 | 8.5 | 68.4 | 3,312 | 15,023 | 313 |
| I | 15.4 | 73.5 | 5.6 | 69.9 | 3,061 | 16,250 | 354 |

Table 3. Analysis of our model on the MOT validation set using different set of components (A) Appearance, (M) Motion, and (I) Interaction. We report the standard MOT metrics.

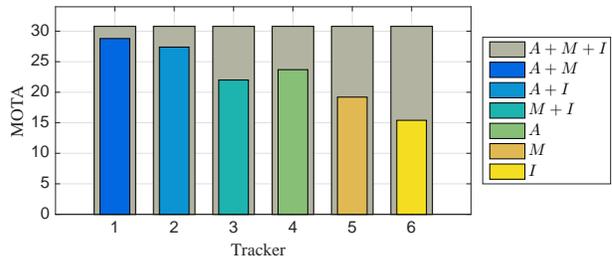

Figure 8. Analysis of our model on the MOT validation set using different set of components (A) Appearance, (M) Motion, and (I) Interaction. We report the MOTA scores.

| Method | Rank 1 | Rank 5 | Rank 10 |
|---|---|---|---|
| FPNN [40] | 19.9 | 49.3 | 64.7 |
| BoW [91] | 23 | 45 | 55.7 |
| ConvNet [2] | 45 | 75.3 | 95 |
| LX [41] | 46.3 | 78.9 | 88.6 |
| MLAPG [42] | 51.2 | 83.6 | 92.1 |
| SS-SVM [88] | 51.2 | 80.8 | 89.6 |
| SI-CI [72] | 52.2 | **84.3** | 92.3 |
| DNS [87] | 54.7 | 84.8 | 94.8 |
| SLSTM [69] | **57.3** | 80.1 | 88.3 |
| **Ours** | 55.9 | 81.7 | **95.1** |

Table 4. Performance comparison our appearance feature extractor with state-of-the-art algorithms for the CUHK03 dataset.

### 4.3. Re-identification Task

For completeness, we report the performance of our appearance cue on re-identification task. We construct a Siamese CNN using the same pre-trained CNN used as our appearance feature extractor in Section 3.2. We train our Siamese CNN on positive and negative samples extracted from two MOT2D and CUHK03 datasets [37, 40]. We extracted more than 500k of positive and negative samples from 2DMOT2015 and CUHK03. In case of MOT2D, we use instances of the same target that occur in different frames for positive pairs, and we use instances of different targets across all frames for negative pairs. Network hyperparameters are chosen by cross validation. The mini-batch size of 64, learning rate of 0.001, sequentially decreased every 2 epochs by a factor 10 (for 20 epochs). We evaluate our appearance model on CUHK03 reidentification benchmark [89]. Table 4 presents our results for Rank 1, Rank 5, and Rank 10 accuracies. Our method achieves 55.9 percent of accuracy for Rank 1 which is competitive against the state-of-the-art method (57.3%). When measuring the re-identification rate for Rank 10, our appearance model outperforms previous methods. This is a crucial indicator for showing that our model can extract meaningful feature representation for re-identification task.

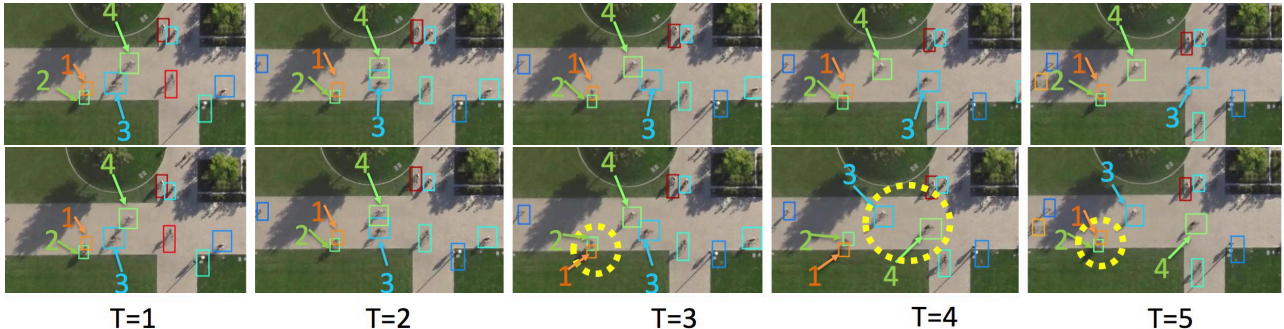

Figure 9. Qualitative results on the Stanford Drone dataset [59]. The first row presents the tracking results of our method whereas the second row presents the results of MDP+SF-mc [59]. The dashed circles illustrate ID switches in previous method.

| Tracker | Tracking Mode | MOTA↑ | MOTP↑ | MT↑ | ML↓ | FP↓ | FN↓ | IDS↓ | Frag↓ | Hz↑ |
|---|---|---|---|---|---|---|---|---|---|---|
| SiameseCNN [36] | Offline | 29.0 | 71.2 | 8.5% | 48.4% | **5,160** | 37,798 | 639 | 1,316 | **52.8** |
| CNNTCM [71] | Offline | 29.6 | 71.8 | 11.2% | 44.0% | 7,786 | 34,733 | 712 | 943 | 1.7 |
| TSMLCDEnew [70] | Offline | 34.3 | 71.7 | 14.0% | 39.4% | 7,869 | 31,908 | 618 | 959 | 6.5 |
| JointMC [27] | Offline | 35.6 | 71.9 | **23.2%** | 39.3% | 10,580 | **28,508** | 457 | 969 | 0.6 |
| TC_ODAL [5] | Online | 15.1 | 70.5 | 3.20% | 55.80% | 12,970 | 38,538 | 637 | 1,716 | 1.7 |
| RMOT [81] | Online | 18.6 | 69.6 | 5.30% | 53.30% | 12,473 | 36,835 | 684 | **1,282** | 7.9 |
| SCEA [79] | Online | 29.1 | 71.1 | 8.9% | 47.3% | 6,060 | 36,912 | 604 | 1,182 | 6.8 |
| MDP [75] | Online | 30.3 | 71.3 | 13.00% | 38.40% | 9,717 | 32,422 | 680 | 1,500 | 1.1 |
| TDAM [80] | Online | 33.0 | **72.8** | 13.3% | 39.1% | 10,064 | **30,617** | **464** | 1,506 | 5.9 |
| **Ours** | Online | **37.6** | 71.7 | **15.8%** | **26.8%** | 7,933 | 29,397 | 1,026 | 2,024 | 1.0 |

Table 5. Tracking performance on the test set of the 2D MOT 2015 Benchmark with public detections.

| Tracker | Tracking Mode | MOTA↑ | MOTP↑ | MT↑ | ML↓ | FP↓ | FN↓ | IDS↓ | Frag↓ | Hz↑ |
|---|---|---|---|---|---|---|---|---|---|---|
| LINF1 [14] | Offline | 41 | 74.8 | 11.60% | 51.30% | 7,896 | 99,224 | 430 | 963 | 1.1 |
| MHT_DAM [31] | Offline | 42.9 | 76.6 | 13.60% | 46.90% | 5,668 | 97,919 | 499 | 659 | 0.8 |
| JMC [78] | Offline | 46.3 | 75.7 | 15.50% | **39.70%** | 6,373 | 90,914 | 657 | 1,114 | 0.8 |
| NOMT [8] | Offline | 46.4 | **76.6** | **18.30%** | 41.40% | 9,753 | **87,565** | 359 | **504** | 2.6 |
| OVBT [6] | Online | 38.4 | 75.4 | 7.50% | 47.30% | 11,517 | 99,463 | 1,321 | 2,140 | 0.3 |
| EAMTT_pub [61] | Online | 38.8 | 75.1 | 7.90% | 49.10% | 8,114 | 102,452 | 965 | 1,657 | **11.8** |
| oICF [30] | Online | 43.2 | 74.3 | 11.30% | 48.50% | 6,651 | 96,515 | **381** | 1,404 | 0.4 |
| **Ours** | Online | **47.2** | 75.8 | **14.0%** | **41.6%** | **2,681** | 92,856 | 774 | 1,675 | 1 |

Table 6. Tracking performance on the test set of the MOT16 Benchmark with public detections.

## 5. Conclusions

We have presented a method that encodes dependencies across multiple cues over a temporal window. Our learned multi-cue representation is used to compute the similarity scores in a tracking framework. We showed that by switching the existing state-of-the-art representation with our proposed one, the tracking performance (measured as MOTA) increases by 20 %. Consequently, our method ranks first in existing benchmarks. As future work, we plan to use our data-driven method to track any social animal such as ants. Their appearance and dynamics are quite different from humans. It will be exciting to learn a representation for such collective behavior and help researchers in biology to get more insights in their field.